\documentclass{article}

\date{} 

\usepackage{enumitem}

\usepackage{url}

\usepackage{hyperref} 

\usepackage{spconf,amsmath,graphicx}
\usepackage{cite}

\title{\emph{SalSi}: A new seismic attribute for salt dome detection}
\name{Muhammad Amir Shafiq, Tariq Alshawi, Zhiling Long and Ghassan AlRegib\thanks{This work is supported by the Center for Energy and Geo Processing (CeGP) at the Georgia Institute of Technology and King Fahd University of Petroleum and Minerals.}}
\address{Center for Energy and Geo Processing (CeGP) at Georgia Tech and KFUPM\\
School of Electrical and Computer Engineering\\
Georgia Institute of Technology, Atlanta, Georgia, 30332-0250\\
\{amirshafiq*,~talshawi,~zhiling.long,~alregib\}@gatech.edu}

\begin{document}

\onecolumn 

\begin{description}[labelindent=1cm,leftmargin=4cm,style=multiline]

\item[\textbf{Citation}]{M. Shafiq, T. Alshawi, Z. Long, and G. AlRegib, ``SalSi: A new seismic attribute for salt dome detection,'' Proceedings of IEEE Intl. Conf. on Acoustics, Speech and Signal Processing (ICASSP), Shanghai, China, Mar. 2016.}
\\
\item[\textbf{DOI}]{\url{https://doi.org/10.1109/ICASSP.2016.7472002}}
\\
\item[\textbf{Review}]{Date of publication: 19 May 2016}
\\
\item[\textbf{Data and Codes}]{\url{https://ghassanalregibdotcom.files.wordpress.com/2016/10/amir_icassp2016_code.zip}}
\\
\item[\textbf{Bib}] {@INPROCEEDINGS\{7472002, \\ 
author=\{M. A. Shafiq and T. Alshawi and Z. Long and G. AlRegib\}, \\ 
booktitle=\{2016 IEEE International Conference on Acoustics, Speech and Signal Processing (ICASSP)\}, \\ 
title=\{SalSi: A new seismic attribute for salt dome detection\}, \\ 
year=\{2016\}, \\ 
pages=\{1876-1880\}, \\ 
doi=\{10.1109/ICASSP.2016.7472002\}, \\ 
ISSN=\{2379-190X\}, \\ 
month=\{March\}\}
} 
\\

\item[\textbf{Copyright}]{\textcopyright 2016 IEEE. Personal use of this material is permitted. Permission from IEEE must be obtained for all other uses, in any current or future media, including reprinting/republishing this material for advertising or promotional purposes, creating new collective works, for resale or redistribution to servers or lists, or reuse of any copyrighted component of this work in other works.}
\\
\item[\textbf{Contact}]{\href{mailto:zhiling.long@gatech.edu}{zhiling.long@gatech.edu}  OR \href{mailto:alregib@gatech.edu}{alregib@gatech.edu}\\ \url{https://ghassanalregib.com/} \\ }
\end{description}

\thispagestyle{empty}
\newpage
\clearpage
\setcounter{page}{1}

\twocolumn

\maketitle

\begin{abstract}
In this paper, we propose a saliency-based attribute, \emph{SalSi}, to detect salt dome bodies within seismic volumes. \emph{SalSi} is based on the saliency theory and modeling of the human vision system (HVS). In this work, we aim to highlight the parts of the seismic volume that receive highest attention from the human interpreter, and based on the salient features of a seismic image, we detect the salt domes. Experimental results show the effectiveness of \emph{SalSi} on the real seismic dataset acquired from the North Sea, F3 block. Subjectively, we have used the ground truth and the output of different salt dome delineation algorithms to validate the results of \emph{SalSi}. For the objective evaluation of results, we have used the receiver operating characteristics (ROC) curves and area under the curves (AUC) to demonstrate \emph{SalSi} is a promising and an effective attribute for seismic interpretation.
\end{abstract}
\section{Introduction}
\label{sec:intro}
The deposition of salt may penetrate into surrounding rock strata such as limestone and shale to form an important diapir structure, a salt dome. Salt, which is impermeable, forms domes that trap hydrocarbon materials including petroleum and natural gas. Therefore, locating salt domes is key to exploring oil and petroleum reservoirs. Experienced interpreters can manually label the boundaries of salt domes by observing and analyzing seismic signals. With the dramatically growing size of acquired seismic data, however, manual labeling of the salt domes is becoming very time consuming and labor intensive. To improve interpretation effectiveness, in recent decades, both industry and academia have used intelligent computer-aided methods to assist the interpretation process. Salt dome delineation, however, poses significant detection and labeling problems because of noise and amplitude variations in seismic data. Fully and semi-automated interpretation algorithms are usually applied to 2D seismic sections to detect an initial boundary of the salt body. Interpreters, based on an initial output, can fix the erroneously detected boundary sections and fine tune the algorithm's parameters to accurately segment salt domes. In this context, having a confidence region around salt dome for delineating boundary and fine tune certain parameters can enhance algorithm's efficiency, reduce computational complexity and speed up the interpretation process.
\begin{figure*}
  \centering
  \includegraphics[width=17.0cm]{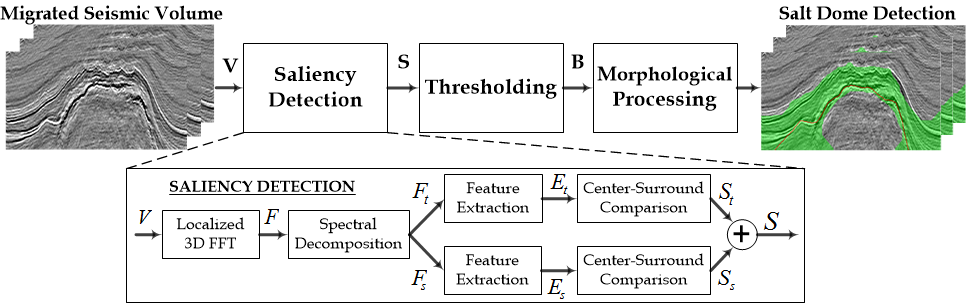}
  \caption{Block diagram of the proposed method}\label{Block_Diag}
\end{figure*}

Researchers, over the last few decades, have proposed several subsurface structures detection methods based on the visual perception of interpreters. In particular, there are several works on the detection of salt domes using graph theory, edge detection, texture, normalized graph cut, active contours and different image processing techniques \cite{lomask2004image, lomask2007application, shi2000normalized, halpert2009seismic, jing2007detecting, aqrawi2011detecting, asjad2015, Shafiq_AC, lse_schlum, winston_LSE, berthelot2013texture, Zhen2015GoT, Shafiq2015GoT}. One of the rarely explored descriptor for seismic interpretation is saliency. Drissi et al. \cite{Drissi_Sal} proposed an algorithm to detect the salient texture features in seismic sections by computing entropy at each pixel using two entropy measures: the Shannon entropy and the generalized cumulative residual entropy. Visual saliency is important to predict the human interpreters attention and highlight the areas of interest in seismic sections. However, the majority of algorithms haven't exploited visual saliency for salt dome detection and seismic interpretation.

Saliency detection aims to highlight salient regions in images and videos by taking into consideration the biological structure of the human vision system (HVS)~\cite{Borji2013}. As a great deal of research in computational cognitive science suggests, HVS has evolved to reduce the amount of the sensory data information gathering stage, also known as the task-free visual search, by focusing on the perceptually salient segments of visual data that conveys the most useful information about the scene~\cite{Borji2015}. Features like color contrast, intensity contrast, flicker, and motion all have been identified as prominent features that help HVS to focus processing resources on important elements in the surrounding environment. One of the most widely accepted theories to characterize saliency and HVS states that localized outliers in both temporal and spatial domains represent novel elements in the environment. The computational aspect of this theory is structured into the center-surround model \cite{Gao08}, in which a saliency map can be obtained by comparing regions in the visual input to its local surrounding in terms of visual feature that have been identified as saliency prominent. Several features and detection algorithms for saliency have been proposed in the literature~\cite{Borji2013, Borji2015}. More recently, a 3D Fast Fourier Transform (FFT)-based saliency detection algorithm for video has been proposed~\cite{Long2015}. This algorithm uses 3D FFT of a non-overlapping window in the spatial and temporal domains of video sequence to compute the spectral energy of the window and compare it with its surrounding regions to construct the saliency map. This method is effective in capturing both temporal and spatial saliency cues in a very fast and compact way. Therefore, we base our research in this paper on the 3D FFT-based saliency detection algorithm.

\begin{figure}
  \centering
  \includegraphics[width=8.0cm]{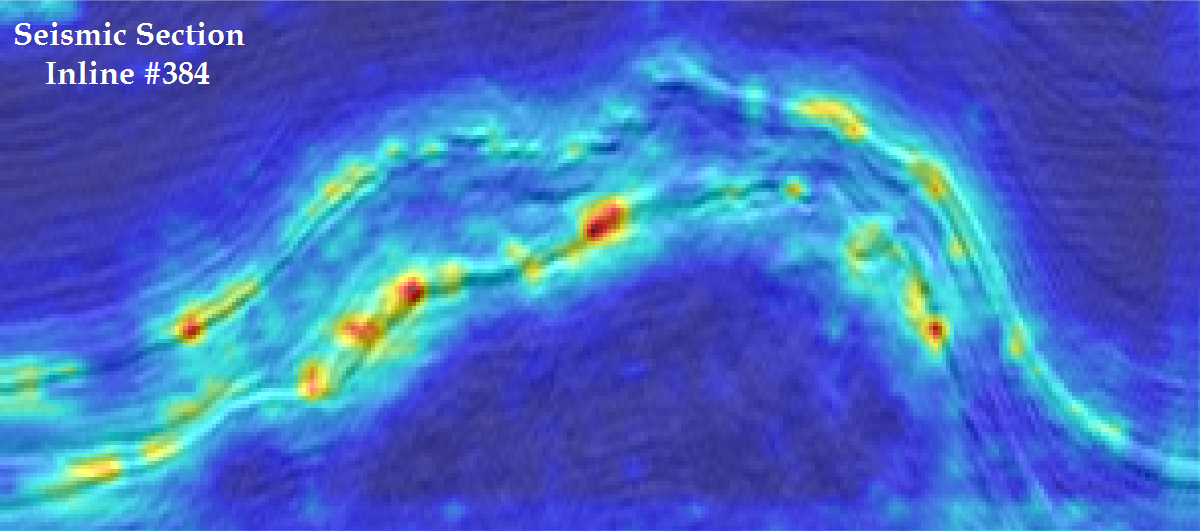}
  \caption{Saliency map superimposed on the seismic section inline \#384}\label{Sal_Map}
\end{figure}

In this paper, we propose a novel saliency-based attribute, \emph{SalSi}. To the best of our knowledge, saliency-based salt dome detection is not reported in the literature. Using \emph{SalSi}, we can process visual stimuli in real-time and perform complex processing procedures faster and more efficiently. The rest of the paper is organized as follows. The proposed salt dome detection method is given in section II. The discussion of experimental results is presented in section III, and finally conclusions are given in section IV.

\begin{figure*}[htb]
\begin{minipage}[b]{.48\linewidth}
  \centering
  \centerline{\includegraphics[width=8.1cm]{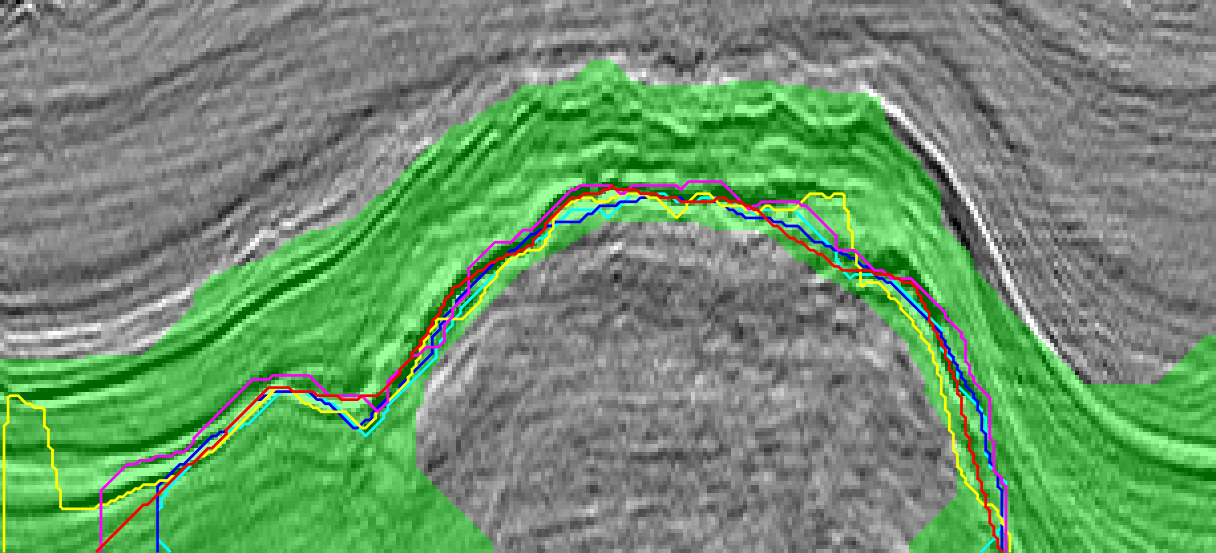}}
  \centerline{(a) Seismic section inline $\#$369}\medskip
\end{minipage}
\hfill
\begin{minipage}[b]{0.48\linewidth}
  \centering
  \centerline{\includegraphics[width=8.1cm]{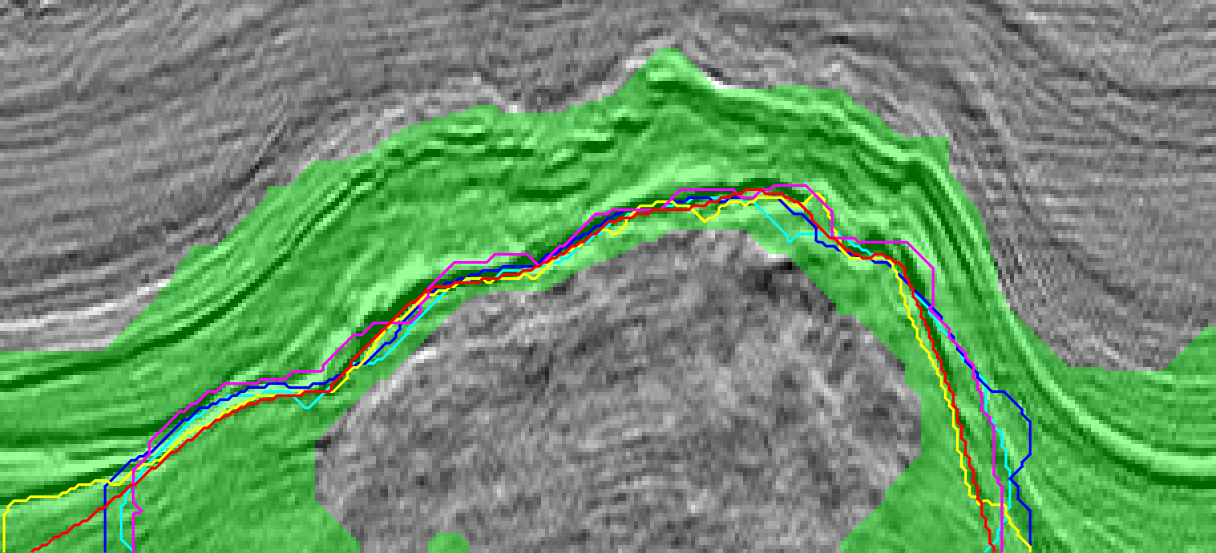}}
  \centerline{(b) Seismic section inline $\#$384}\medskip
\end{minipage}
\begin{minipage}[b]{.48\linewidth}
  \centering
  \centerline{\includegraphics[width=8.1cm]{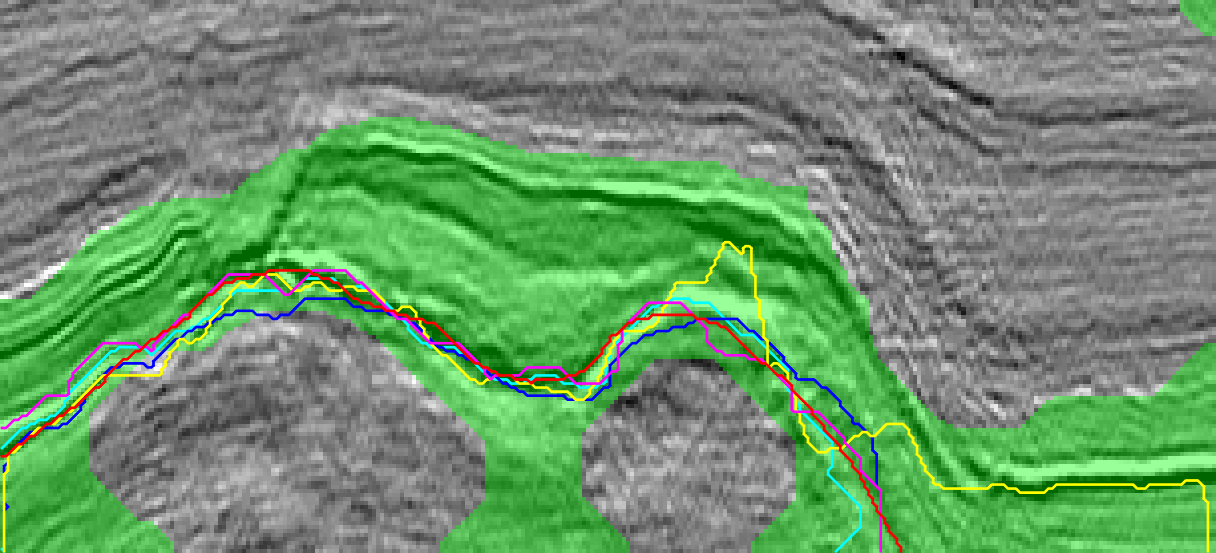}}
  \centerline{(c) Seismic section inline $\#$429}\medskip
\end{minipage}
\hfill
\begin{minipage}[b]{0.48\linewidth}
  \centering
  \centerline{\includegraphics[width=8.1cm]{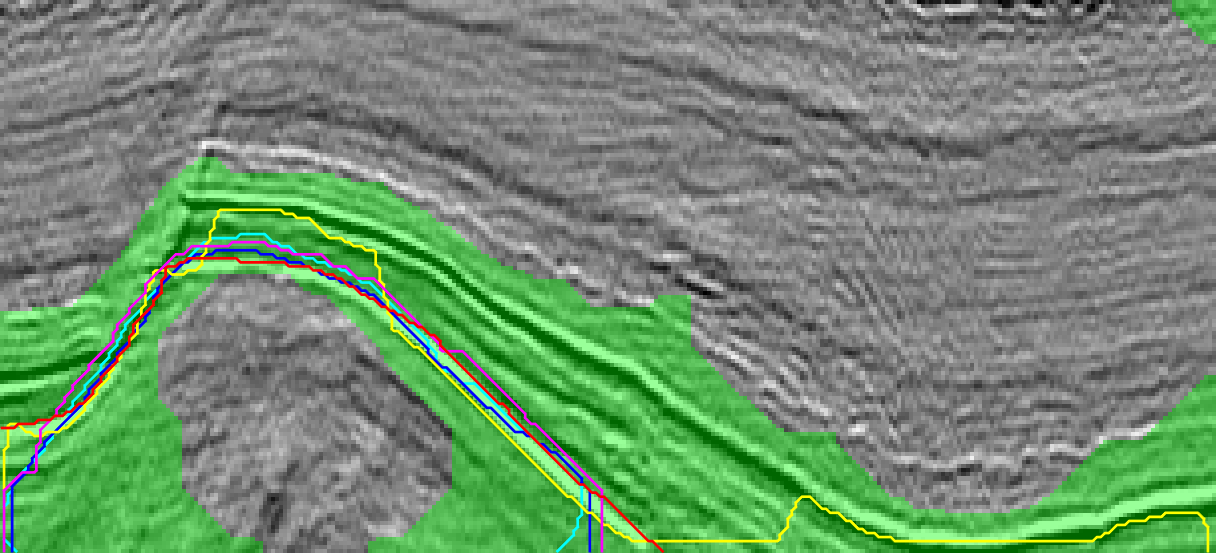}}
  \centerline{(d) Seismic section inline $\#$459}\medskip
\end{minipage}
\caption{Experimental results: \emph{SalSi} output superimposed on different seismic sections. Magenta: Aqrawi et al.~\cite{aqrawi2011detecting}, Yellow: Berthelot et al.~\cite{berthelot2013texture}, Cyan: Zhen et al.~\cite{Zhen2015GoT}, Blue: Shafiq et al.~\cite{Shafiq2015GoT}, Green: \emph{SalSi} Output, Red: Ground Truth.}
\label{SalSeisop}
\end{figure*}

\section{Proposed Method for Salt Dome Detection}
\label{sec:proposed}

An overall block diagram of \emph{SalSi} is shown in Fig.~\ref{Block_Diag}. For the application under consideration, the most salient part of a seismic image is salt dome boundary. Thus, given a 3D seismic data volume  $\boldsymbol V$ of size $M \times N \times K$, where $M$ represents time depth, $N$ represents crosslines and $K$ represents inlines, we compute saliency using the 3D FFT-based algorithm proposed in~\cite{Long2015}. The 3D FFT-based saliency algorithm is fast, and obtains saliency maps with adjustable resolution, that allows better salient objects segmentation. The 3D FFT-based algorithm is computationally inexpensive and requires very few parameters as compared to other visual saliency algorithms, which make it advantageous for seismic applications. The block diagram of 3D FFT-based saliency detection algorithm is also shown in Fig.~\ref{Block_Diag}.

To perform the saliency detection, first we calculate the 3D FFT spectrum $\boldsymbol F$ in a local area using (\ref{eqn:3dfft}), and decompose $\boldsymbol F$ into a temporal-change-related component $\boldsymbol F_t$ and a spatial-change-related component $\boldsymbol F_s$.
\begin{equation} \label{eqn:3dfft}
F[u,v,w]=\frac{1}{L^3}\sum\limits_{m=0}^{L-1}\sum\limits_{n=0}^{L-1}\sum\limits_{k=0}^{L-1}f[m,n,k]e^{-2\pi i\left(mu+nv+kw\right)/L},
\end{equation}
\begin{equation}
F_t[u,v,w]=F[u,v,w] \times \frac{w}{\sqrt[]{u^2+v^2+w^2}},
\end{equation}
\begin{equation}
F_s[u,v,w]=F[u,v,w] \times \frac{\sqrt[]{u^2+v^2}}{\sqrt[]{u^2+v^2+w^2}},
\end{equation}
\noindent where $[m,n,k]$ and $[u,v,w]$ represent the coordinates in the space and frequency domains, respectively, $L$ defines the size of the local data cube, and $f[m,n,k]$ is the seismic image or section. Subsequently, spectral energies $\boldsymbol E_t$ and $\boldsymbol E_s$ are extracted from $\boldsymbol F_t$ and $\boldsymbol F_s$, respectively, as features. Applying the center-surround model, two saliency maps $\boldsymbol S_t$ and $\boldsymbol S_s$ can be constructed using $\boldsymbol E_t$ and $\boldsymbol E_s$ as
\begin{align}\label{eqn:S_x}
S_x[m,n,k]=& \frac{1}{Q} \sum_{i_0,j_0,r_0} | E_x[m,n,k] \nonumber\\
& - E_x[m+i_0,n+j_0,k+r_0] |,
\end{align}
\noindent where $i_0$, $j_0$, $r_0$ are chosen such that point $[m+i_0,n+j_0,k+r_0]$ is in the immediate neighborhood of point $[m,n,k]$, such as within a $3 \times 3 \times 3$ window centered at $[m,n,k]$, $Q$ is the total number of points included in the summation, $\boldsymbol S_x$ represents $\boldsymbol S_t$ or $\boldsymbol S_s$, and $\boldsymbol E_x$ represents $\boldsymbol E_t$ or $\boldsymbol E_s$. The final saliency map $\boldsymbol S$ is obtained by averaging $\boldsymbol S_t$ and $\boldsymbol S_s$, and is of same size as of $\boldsymbol V$.
\begin{equation} \label{eqn:S}
S[m,n,k]=0.5 \times S_t[m,n,k] + 0.5 \times S_s[m,n,k].
\end{equation}

The second step of \emph{SalSi} is to threshold the saliency map, $\boldsymbol S$ as in (\ref{eqn:thres}).

\begin{equation}
\label{eqn:thres}
\mathbf{B}[m,n,k]=
\left\{
\begin{aligned}
&1\quad \mathbf{S}[m,n,k]\geq \mathbf{T}\\
&0\quad \mbox{Otherwise}
\end{aligned}
\right.
,
\end{equation}
\noindent where $\mathbf{B}$ represents the binary volume and white regions in $\mathbf{B}$ will likely contain salt dome boundaries. We calculate the threshold value, $\mathbf{T}$, using Otsu's method \cite{Otsu_method}. In contrast to the non-salt regions, salt dome boundaries commonly have higher $\boldsymbol S$ values. Therefore, we assume that the histogram of volume $\boldsymbol S$ follows a bi-modal distribution shape. To optimally divide all points into two classes, we determine the threshold $T$ by minimizing the intra-class variance as follows:
\begin{equation}
\label{equ:otsu}
\arg\underset{T}{\min}\left\{\sigma_{1}^2(T)\sum_{i=0}^{T-1}p(i)+\sigma_{2}^2(T)\sum_{i=T}^{H}p(i)\right\},
\end{equation}
\noindent where $H$ is the number of quantized gray-levels of $\boldsymbol S$, and $p(i)$, $i=0,\cdots,H-1$, represents the probability of points with gray value $i$. In addition, $\sigma_1^2$ and $\sigma_2^2$ define the individual class variances, which can be calculated as follows:
\begin{equation}
\label{equ:var}
\left\{
\begin{aligned}
&\sigma_1^2=\sum\limits_{i=0}^{T-1}\left[i-\sum\limits_{i=0}^{T-1}\frac{iP(i)}{P_1}\right]^2\frac{P(i)}{P_1},\quad P_1=\sum\limits_{i=0}^{T-1}P(i)\\
&\sigma_2^2=\sum\limits_{i=T}^{H-1}\left[i-\sum\limits_{i=T}^{H-1}\frac{iP(i)}{P_2}\right]^2\frac{P(i)}{P_2},\quad P_2=\sum\limits_{i=T}^{H-1}P(i)\\
\end{aligned}
\right.
\end{equation}
Therefore, on the basis of (\ref{equ:otsu}), we can adaptively identify threshold $T$ by exhaustively searching between $0$ and $H-1$.

Salt domes are complicated structures  and after applying the threshold $\boldsymbol T$ on $\boldsymbol S$, it is inevitable that the binary volume $\boldsymbol B$ contains noisy or disconnected boundary regions. In order to process noisy and disconnected boundary regions, finally we apply morphological closing operation as a post processing step to $\boldsymbol B$ to ensure salt body is closed. In morphological closing operation, which is comprised of dilation followed by erosion, we have used circular disk of radius ten as structuring element $\boldsymbol {Se}$. As a result of morphological operation, \emph{SalSi} will generate a salient map of a seismic image that highlights the salt dome boundary.

\begin{figure}
  \centering
  \includegraphics[width=8.0cm]{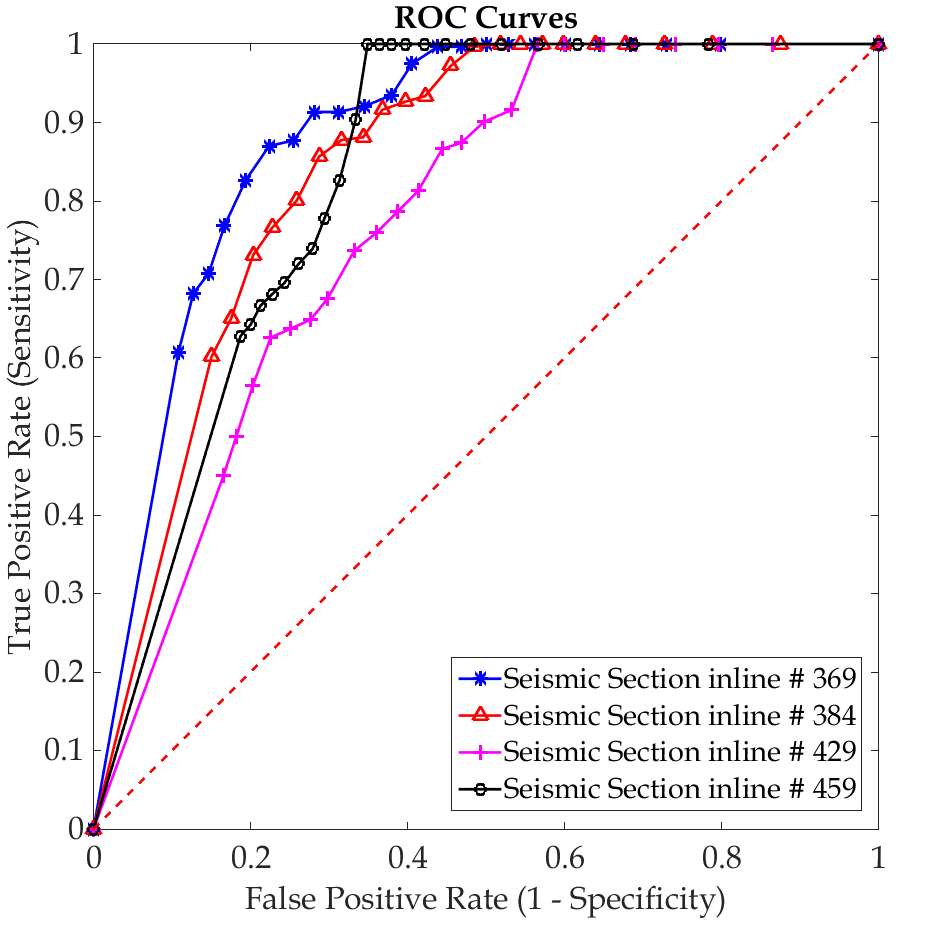}
  \caption{ROC curves}\label{Perf_curves}
\end{figure}

\section{Experimental Results}
\label{sec:results}
In this section, we present the effectiveness of \emph{SalSi} for salt dome detection. We have used the real seismic dataset acquired from the Netherland's offshore, $F3$ block in the North Sea, whose size is $24$ x $16$ $km^2$ \cite{F3_data}. The seismic volume that contains the salt dome structure has an inline number ranging from $\#151$ to $\#501$, a crossline number ranging from $\#401$ to $\#701$, and a time direction ranging from $1,300ms$ to $1,848ms$ sampled every $4ms$.

A typical seismic section's saliency map superimposed on original image is shown in Fig.~\ref{Sal_Map}. It can be observed from Fig.~\ref{Sal_Map} that the saliency map depicts boundary at the salt dome perimeter. The output of \emph{SalSi}, and the results of different salt dome delineation algorithms on seismic section inlines \#369, \#384, \#429 and \#459 are shown in Fig.~\ref{SalSeisop}, with the ground truth manually labeled in red. The magenta, yellow, cyan and blue lines represent the boundaries detected by \cite{aqrawi2011detecting}, \cite{berthelot2013texture}, \cite{Zhen2015GoT} and \cite{Shafiq2015GoT}, respectively, whereas the green area represents \emph{SalSi} output. Subjectively, it can be observed that the boundaries detected by all aforementioned algorithms lie within \emph{SalSi} output, which make it suitable for algorithms initialization, tracking boundaries within seismic volumes, and reducing time computation.

To objectively evaluate the effectiveness of \emph{SalSi} with different thresholds, we calculated the receiver operating characteristics (ROC) curves, and the area under the curve (AUC) of different seismic section inlines. We used multi-level threshold \cite{Liao01afast} and measured \emph{SalSi} performance using two important statistical measures, sensitivity and specificity, usually used in binary classification. Sensitivity, which is also called true positive rate (TPR) measures the voxels that are correctly identified using \emph{SalSi}. On the other hand, specificity, also known as true negative rate measures the proportion of voxels that don't belong to the salt dome boundary and are correctly identified as such. Fall out rate also termed at false positive rate or false alarm can be calculated as (1-specificity). ROC curves, obtained by plotting sensitivity against fall out rate, are shown in Fig.~\ref{Perf_curves}. These curves illustrates the performance of \emph{SalSi} as a binary classifier when threshold is varied between a certain range. A good detection method has ROC curve over the random selection method, depicted by dashed red line in Fig.~\ref{Perf_curves}, and it can be observed that all ROC curves of \emph{SalSi} are above random selection method. An optimum threshold can also be obtained by averaging all ROC curves and comparing TPR with FPR. A good threshold value will generate a saliency map with maximum sensitivity and minimum fall out rate. We also calculated the AUC values from the ROC curves, and results are presented in Table.~1. The AUC values closer to one indicate that the detection method is very good and Table.~1 demonstrates that \emph{SalSi} performs very well on different seismic section inlines and has mean AUC of 0.8324. Therefore, experimental results presented in this paper show a promising future of \emph{SalSi} for salt dome detection and very good potential for seismic interpretation.

\begin{table}[t]
\begin{center}
\caption{The AUC values of the ROC curves}\label{table1}
\medskip

\begin{tabular}{|c|c|c|}
\hline
\textbf{Sr. $\#$}  & \textbf{Seismic Section} & \textbf{~~~AUC~~~} \\
\hline
1 & Inline $\#$ 369 & 0.8803 \\
\hline
2 & Inline $\#$ 384 & 0.8434 \\
\hline
3 & Inline $\#$ 429 & 0.7740 \\
\hline
4 & Inline $\#$ 459 & 0.8317 \\
\hline
\end{tabular}
\end{center}
\end{table}

\section{Conclusion}
\label{sec:conclusion}
In this paper, we have proposed a novel saliency-based attribute, \emph{SalSi}, for salt dome detection from seismic volumes. \emph{SalSi} can be used to initialize algorithms and to reduce computational complexity of algorithms. Therefore, \emph{SalSi} is also suitable for many seismic interpretation applications such as salt dome delineation, tracking salt domes in seismic volume, seismic retrieval and labeling, etc. \emph{SalSi}, originally designed to detect salt domes can be modified to capture chaotic horizons and faults as well from seismic volumes. The experimental results show that \emph{SalSi} can accurately locate salt domes within seismic volumes and enhance the efficiency of interpretation process. The AUC and ROC curves demonstrate that \emph{SalSi} performs very well even with the different threshold values, which validate \emph{SalSi} as an effective attribute for seismic interpretation. Initial results of \emph{SalSi} presented in this paper are very encouraging having various applications in seismic interpretation.

\bibliographystyle{IEEEbib}
\bibliography{main}

\end{document}